\PassOptionsToPackage{usenames,dvipsnames}{xcolor}
\documentclass[9pt, twocolumn, twoside, lineno]{article}
\usepackage{setspace}

\usepackage[font=small,labelfont=bf]{caption}
\usepackage{siunitx}
\usepackage{xcolor}
\usepackage[utf8]{inputenc}
\usepackage{tikz-dependency}
\usepackage{tikz}
\usepackage{relsize}
\usepackage{anyfontsize}
\usepackage[normalem]{ulem}
\usetikzlibrary{calc, shadows, shapes,arrows, decorations.pathmorphing, arrows.meta,bending,matrix,positioning}
\tikzstyle{naveqs} = [text width=40mm, fill=ForestGreen, minimum height=20mm, rounded corners]
\tikzstyle{ann} = [above, text width=5em]
\usepackage{anyfontsize}
\usepackage{authblk}
\usepackage{cite}
\usepackage[colorlinks,allcolors=blue]{hyperref}
\usepackage{titlesec}
\usepackage{geometry}
\renewcommand{\thesubsection}{\Alph{subsection}}
\titleformat{\subsection}[runin]{\normalfont\fontsize{9}{3}\bfseries}{\thesubsection}{1em}{}
\titleformat{\subsubsection}[runin]{\normalfont\fontsize{8}{3}\bfseries}{\thesubsubsection}{1em}{}
\usepackage{varwidth}
\usepackage{amsmath}
\usepackage{amssymb}
\usepackage{amsfonts}
\usepackage{mathrsfs}
\usepackage{mathtools}

\usepackage{amsthm}
\usepackage{laterref}

\newcommand{\state}{x}

\newcommand{\stateset}{\mathcal{X}}
\newcommand{\statevalueset}{\mathbb{R}^{d_x}}
\newcommand{\control}{a}

\newcommand{\controlset}{\mathcal{A}}
\newcommand{\controlvalueset}{\mathbb{R}^{d_a}}

\newcommand{\Ent}{H}
\newcommand{\I}{I}
\newcommand{\C}{\mathcal{C}}
\newcommand{\defineas}{:=}

\newcommand{\pendangle}{\theta}
\newcommand{\observedpendangle}{\tilde{\theta}}
\newcommand{\observationnoise}{\tilde{\eta}_{\text{obs}}}
\newcommand{\angularvelocity}{\dot{\pendangle}}

\newcommand{\torque}{a}

\newcommand{\pendulumlength}{l}
\newcommand{\transpose}{\dagger}

\newcommand{\totaltimeint}{{T_e}}
\newcommand{\actiontimeint}{{T_a}}
\newcommand{\totaltime}{t_e}
\newcommand{\actiontime}{t_a}
\newcommand{\timegapint}{{\Delta T}} % not ideal, as its semantics is
\newcommand{\timeint}{0}        % current time in index form
\newcommand{\timereal}{t}
\newcommand{\deltatimeint}{1}

\newcommand{\deltat}{\Delta t}        %
\newcommand{\timeindex}{k}

\newcommand{\channelpower}{\sigma}

\providecommand{\keywords}[1]
{
  \small	
  \textbf{\textit{Keywords---}} #1
}

\makeatletter
\renewcommand\AB@affilsepx{, \protect\Affilfont}
\renewcommand\Affilfont{\fontsize{9}{10.8}\itshape}

\title{Intrinsic Motivation in Dynamical Control Systems}

\author[a]{Stas Tiomkin}
\author[b,c,d]{Ilya Nemenman} 
\author[e]{Daniel Polani}
\author[f,g]{Naftali Tishby\thanks{\it Prof.\ Naftali Tishby passed away when this work was in development. This project began under his leadership when Stas Tiomkin was a PhD student in his group. The rest of the authors agree that he should be a senior author on this manuscript, but his consent for this was not obtained.}}

\affil[a]{Computer Engineering Department, Charles W. Davidson College of Engineering San Jose State University, CA, 95192}
\affil[b]{Department of Physics}
\affil[c]{Department of Biology}
\affil[d]{Initiative in Theory and Modeling of Living Systems, Emory University, Atlanta, GA 30322, USA }
\affil[e]{Adaptive Systems Research Group, University of Hertfordshire, Hatfield, UK}
\affil[f]{The Rachel and Selim Benin School of Computer Science and Engineering}
\affil[g]{Edmond and Lilly Safra Center for Brain Sciences (ELSC), Hebrew University of Jerusalem, 96906 Israel}

\date{}

\begin{document}
\maketitle
\setstretch{1.04}

\begin{abstract}
Biological systems often choose actions without an explicit reward signal, a phenomenon known as intrinsic motivation. The computational principles underlying this behavior remain poorly understood. In this study, we investigate an information-theoretic approach to intrinsic motivation, based on maximizing an agent's empowerment (the mutual information between its past actions and future states). We show that this approach generalizes previous attempts to formalize intrinsic motivation, and we provide a computationally efficient algorithm for computing the necessary quantities. We test our approach on several benchmark control problems, and we explain its success in guiding intrinsically motivated behaviors by relating our information-theoretic control function to fundamental properties of the dynamical system representing the combined agent-environment system. This opens the door for designing practical artificial, intrinsically motivated controllers and for linking animal behaviors to their dynamical properties.
\end{abstract}

\vspace{0.5cm}

\keywords{information capacity $|$ sensitivity gain $|$ stabilization $|$ predictive information} 

\vspace{0.5cm}
%\subsubsection*{Introduction}
\section*{Introduction}
Living organisms are able to generate behaviors that solve novel challenges without prior experience. Can this ability be explained by a single, generic mechanism? One proposal is that novel, useful behaviors can be generated through {\em intrinsic motivation}  \cite{oudeyer2009intrinsic}, which is defined informally as a set of computational algorithms that are derived directly from the intrinsic properties of the organism-environment dynamics and not specifically learned.

Increasingly, there is a move away from reinforcement learning and its extrinsically specified reward structure  \cite{sutton2018reinforcement,doya2000reinforcement} in the theory and practice of artificial agents, robots, and machine learning more generally \cite{mohamed2015variational,gregor2016variational,baumli2021relative,kwon2020variational,sharma2019dynamics,sharma2020emergent,eysenbach2018diversity,houthooft2016vime, achiam2018variational,choi2021variational,salge2014empowerment,klyubin2005empowerment,jung2011empowerment,klyubin08:_keep_your_option_open,zhao2020efficient, zhao2019dynamical,wissner2013causal}. A specific class of such intrinsic motivation algorithms for artificial systems is known as {\em empowerment maximization}. It proposes that agents should maximize the mutual information  \cite{Cover}  between their potential actions and a subsequent future state of the world \cite{Empowerment1}. This corresponds to maximizing the diversity of future world states achievable as a result of the chosen actions, potentiating a broader set of behavior options in the future. Intrinsically motivated synthetic agents develop behaviors that are atypical for inanimate engineered systems and often resemble those of simple living systems. Interestingly, potentiating future actions is also a key part of the success of modern reward-based training algorithms \cite{sharma2019dynamics,EmpBerkly1,Du2020AvEAV}.

Despite the successes of empowerment maximization, it remains unclear how well it can be used as a general intrinsic motivation principle. There are many different versions of intrinsic motivation related to empowerment, and their relation to each other is unknown \cite{wissner2013causal, EmpBerkly1, salge2017empowerment}. Additionally, most work on empowerment maximization has relied on simulational case studies and ad hoc approximations, and analytical results are scarce. In order to gain insight, it is important to link empowerment  to other, better-understood characterizations of the systems in question. Finally, calculating the mutual information between two interlinked processes in the general case is a challenging task \cite{bialek2001predictability, holmes}, which has so far limited the use of empowerment maximization to simple cases.

In this work, we unify different versions of intrinsic motivation related to the empowerment maximization paradigm. Here our main contribution is in showing analytically that empowerment-like quantities are linked to the sensitivity of the agent-environment dynamics to the agent's actions. This connects  empowerment maximization to well-understood properties of dynamical systems. Since highly sensitive regions of the dynamics potentiate many diverse  future behaviors, the connection to dynamical systems also explains why empowerment-based intrinsic motivations succeed in generating behaviors that resemble those of living systems.

The analytical results allow us to develop a practical computational algorithm for calculating empowerment for complex scenarios in the continuous time limit, which is the second major contribution of the paper. We apply the algorithm to standard benchmarks used in intrinsic motivation research \cite{Empowerment9,salge2014empowerment,anonymous2021efficient}. Specifically, a controller based on the efficient calculation of empowerment manages to balance an inverted pendula without extrinsic rewards. This opens the door for designing complex robotic intrinsically motivated agents with systematically computed --- rather than heuristically estimated --- empowerment.

\begin{figure*}[t!]
\begin{tikzpicture}[every text node part/.style={align=center}, scale=1.0]
    % These are needed to remove the vertical space around above and below
    % the align and flalign environments
    \setlength{\abovedisplayskip}{0pt}
    \setlength{\belowdisplayskip}{0pt}

    \node (example) at (0,0) {
        \begin{minipage}{\textwidth}
            \begin{align*}
         \underset{}{\mathcal{C}^{\totaltimeint, \actiontimeint,
              \timegapint}
              \bigl(x_{\timeint}\equiv x(\timereal)\bigr)}=\underset{p(\vec{a}\mid x_\timeint)}{\max}\;\mathcal{I}[\underbrace{\{X_{\totaltimeint}, X_{\totaltimeint-\deltatimeint}, \dots, X_{\actiontimeint+ \timegapint}\}}_{\vec{X}-\mbox{future states}};\underbrace{\{A_{\actiontimeint-\deltatimeint}, A_{\actiontimeint-2}, \dots, A_{\timeint}\}}_{\vec{A}-\mbox{possible actions}}\mid x_\timeint\equiv x(\timereal)]
            \end{align*}
        \end{minipage}
        };
    \node(eq1) at (-5.0, -1.75) {\textbf{Empowerment}\\$\actiontimeint=\totaltimeint, \timegapint=0$};
    \node(eq2) at (-2.75, -1.75) {\textbf{$\cdots$}};
    \node(eq3) at (+0.0, -1.75) {\textbf{controlled Lyapunov expt.}\\$\actiontimeint=1, \timegapint=\totaltimeint-\deltatimeint$};
    \node(eq4) at (+2.75, -1.75) {\textbf{$\cdots$}};
    \node(eq5) at (+5.0, -1.75) {\textbf{kicked CEF}\\$\actiontimeint=1, \timegapint=0$};
    \draw [->, very thin] (example.south) -- (eq1.north);
    \draw [->, very thin, dashed] (example.south) -- (eq2.north);
    \draw [->, very thin] (example.south) -- (eq3.north);
    \draw [->, very thin, dashed] (example.south) -- (eq4.north);
    \draw [->, very thin] (example.south) -- (eq5.north);
\end{tikzpicture}
\caption{\label{fig:scheme}Unified view on information theoretical intrinsic motivation,  for a discretized process sequence.  Starting at time $x_\timeint$ (i.e.\ $x(\timereal)$), potential  actions are  applied for $\actiontimeint$ times, following that, after waiting for  $\timegapint$ time steps, the future system trajectory is considered until    $\totaltimeint$. A controlled Lyapunov exponent is a Lyapunov   exponent, but only in directions controlled by the agent, cf., \eqref{eq:LyapExp}. ``Kicked CEF'' refers to a variant of Causal   Entropic Forcing \cite{CEF}, with the addition that an action kicks the system  at  the beginning of a trajectory. For more details see {\em Generalized Empowerment}.}
\end{figure*}

% \significancestatement{%
 % \subsubsection*{Contribution} We investigate an approach to intrinsic motivation in decision-making based on information-theoretic principles. We show that our method, which involves selecting actions that improve the controller's ability to affect the future, leads to behaviors that mimic those of biological organisms in various situations without the need for explicit reward signals. We also provide an efficient algorithm for computing various versions of this intrinsic motivation, using the small noise/small control approximation to the dynamics. This algorithm is local, requiring only knowing the dynamics in the vicinity of the current state of the system. Our approach opens up opportunities for scaling to more complex scenarios, and provides insight into how intrinsic motivations can emerge from the underlying dynamics of the system.
% }
\newpage
\section*{Results}

\subsection{Preliminaries}

\subsubsection*{Notation} 

We consider an agent that takes on states $x(t) \in \stateset := \statevalueset$, evolving in time under the dynamics $f$ with (small) stochastic perturbations $\eta(t) \in \statevalueset$. Via its (small) actions, $a(t) \in \controlset := \controlvalueset$ filtered through the control gain $g$, the agent can affect the dynamics of the system:
\begin{align}
  d\state(t) = f(\state(t))dt + g(\state(t))da(t) + d\eta\;\label{eq:FullDynCont}.
\end{align}
Here $d\eta$ denotes the system noise, modeled as a Wiener process. The agent's actions  $a(t)$ are modeled by a stochastic control process with variance $\sigma^2_t$ controlled by the agent and with a  mean of zero. This models potential effect of actions centered around the null action. 

To compute  various  quantities of interest,  we will consider a discretized versions of this system, for which we adopt a modified notation. To distinguish it from the continuous version, we replace the continuous time in parentheses by an integer index, $x_\timeindex := x(t+\timeindex\cdot \deltat)$. Here  $\deltat$ denotes the physical time step, and we adopted the convention that $x_0=x(t)$, so that the index corresponding to the current physical time, $t$, is chosen as 0. We will consider trajectories of a fixed  duration, and the agent will apply actions over a part of that trajectory. We denote by $T_e$ the time index of the very last state of the trajectory,  which we also refer to as the {\em time horizon}. We further use $T_a$ to denote the (discretized) duration of the action sequence. Then state, control and perturbation trajectories at finite equidistant times, $\{t+\timeindex\cdot \deltat\}_{\timeindex=0}^{T}$,
are denoted by
$\state_0^\totaltimeint \equiv
\{\state_\timeindex\}_{\timeindex=0}^\totaltimeint$,
$\control_0^\actiontimeint \equiv
\{\control_\timeindex\}_{\timeindex=0}^\actiontimeint$, and
$\eta_0^\totaltimeint \equiv
\{\eta_\timeindex\}_{\timeindex=0}^\totaltimeint$,
respectively. For consistency with the control theory literature, we write a trajectory in the reverse order, e.g., $x_0^\totaltimeint=(x_\totaltimeint, \dots x_0)$. When we wish to emphasize the continuous nature of the underlying process, we will write $\totaltime \equiv t+\totaltimeint \cdot \deltat$ and $\actiontime \equiv t+\actiontimeint \cdot \deltat$ for explicitly continuous times.

\subsubsection*{Reinforcement Learning vs.\ Intrinsic Motivation} 

To elicit a desired behavior in an agent, one typically uses reinforcement learning (RL).  RL is task-specific, and an agent needs an {\em  extrinsic} feedback about its performance from a {\em reward} function to learn  the behavior. The precise construction of this reward function is critical to achieve a desired performance in a short training time \cite{sutton2018reinforcement}. Some of the complications include a significant degree of arbitrariness when choosing amongst reward functions with equivalent performance \cite{ng1999policy} and the difficulty of translating an often vague desired behavior into a concrete reward function. Furthermore, complex behaviors consist of combinations of shorter sequences.  Designing a reward function capable of partitioning the solution into such parts and hence learning it in a realistic time is hard  \cite{dayan1993feudal}. 

In contrast to this, in living systems, acquisition of skills often starts with task-unspecific learning. This endows organisms with {\em potentiating} skills, which are not rewarding on their own. This is then followed by task-oriented specialization, which combines task-unspecific behaviors into complex and explicitly rewarding tasks \cite{oudeyer2009intrinsic,hrl2016}. While specific tasks are often refined with the help of an extrinsic reinforcement, the potentiating tasks usually are intrinsically motivated \cite{sharma2020emergent}.

\subsubsection*{Empowerment} 
The type of intrinsic motivation we focus on is {\em empowerment}. Empowerment is based on information-theoretic quantities \cite{pathakICLR19largescale,pathakICMl17curiosity,EmpBerkly1,Volks,mohamed2015variational,Empowerment2,Empowerment3,Empowerment12,CEF,charlesworth2019intrinsically}. It defines a pseudo-utility function on the state space, based on the system dynamics only, without resorting to a reward. Formally, we express the dynamics of the system by the conditional probability distribution  $p(x_{\totaltimeint}\mid a_0^{\totaltimeint-1}, \state_0)$ of the resulting state when one starts in a state $\state_0$ and subsequently carries out an action sequence $a_0^{\totaltimeint-1}$. Then the empowerment $\C(\state_0)$ is a function of the starting state, $\state_0$. It is given by the maximally achievable  mutual information (the channel capacity \cite{Cover}) between the control action sequence of length $\totaltimeint$ and the final state when starting in the state $\state_0$:
\begin{align}
  \label{eq:empowerment-definition-formal}
    \C(\state_0) &\defineas \underset{p(\control_0^{\totaltimeint-1}\mid \state_0)}{\max}\;\;\I(X_{\totaltimeint};A_{0}^{\totaltimeint-1}| \state_0).
\end{align}
Here $p(\cdot)$ denotes a probability density or a probability distribution function,  and $I$ is the mutual information \cite{Cover}
\begin{equation}
   \I(X_{\totaltimeint};A_{0}^{\totaltimeint-1}|\state_0)=\Ent(X_\totaltimeint|\state_0)-\Ent(X_\totaltimeint\mid A_{0}^{\totaltimeint-1}\state_0)\label{eq:MI_EMP_DEF}.
\end{equation}
$\Ent$ is the entropy, and conditioning an entropy on a random variable means the entropy of the  conditional distribution, averaged over the conditioning variable. The empowerment $\C(\state_0)$ depends on both the state, $x_0$, and the time horizon, $\totaltimeint$. However, for notational convenience, we omit all parameters from the notation except for the dependency on $\state_0$.

Locally maximizing empowerment (e.g., by following its gradient over $\state_0$) guides an agent to perform actions atypical within the natural dynamics of the system. Indeed, since empowerment measures  the diversity of achievable future states, maximizing it increases this diversity (``empowers'' the agent -- hence the name). Thus it is expected to be particularly useful for learning potentiating tasks  \cite{sharma2020emergent}. Crucially, empowerment quantifies the relation between the final state and the \emph{intentional} control, rather than the diversity of states due to the stochasticity of the system. In particular, it is not just the entropy of a passive diffusion process in the state variables, but of the subprocess that the agent can actively generate. Furthermore, it quantifies diversity due to \emph{potential} future action sequences, which are not then necessarily carried out.

Empowerment is typically used in the form of the  {\em empowerment maximization principle} \cite{klyubin08:_keep_your_option_open}, treats  $\C(\state_0)$ as a pseudo-utility function. At each time step, an agent chooses an action to greedily optimize its empowerment at the next time step. That is, the agent climbs up in its empowerment landscape, eventually achieving a local maximum of $\C$:
\begin{align}
  \control^{*}
  \bigl(
  \state(t)
  \bigr)&=\underset{\control\in\controlset}{\mbox{argmax}}\;\mathbb{E}_{\eta}\bigl[\C
          \bigl(
          f(\state(t)) + g(\state(t))\control \Delta {t}' + d\eta
          \bigr)\bigr]\;.
\label{EmpOptAction}
\end{align}
Here $\controlset$ is the set of permitted actions, $\Delta {t}'$ is a small time step used to simulate the actual behavior of the system (and which is selected independently from the time step $\Delta t$ used to discretize \eqref{eq:FullDynCont}). An empowerment-maximizing agent  generates its behavior by repeating this action selection procedure for each decision step it takes.

Crucially, no general analytical solutions or efficient algorithms for numerical estimation of  empowerment for arbitrary dynamical systems are known, limiting  adoption of the empowerment maximization principle. Our goal  is to provide a method to calculate it under specific  approximations. 

\subsection{Empowerment in dynamical systems}
\subsubsection*{The linear response approximation}
To relate empowerment to traditional quantities used to describe dynamical systems, we assume that the control signal $a$ in \eqref{eq:FullDynCont} is small. This is true in some of the most interesting cases, where the challenge is to solve a problem with only {\em weak controls}  that cannot easily ``force'' a solution. Under this assumption,  \eqref{eq:FullDynCont} is approximated by a linear time-variant dynamics around the trajectories of the autonomous dynamics (i.e., for $a=0$).  To proceed, we now introduce the following notation. We define $\bar{x}_s$ as the $s$-th step of the trajectory in the discretized deterministic approximation of the  dynamics \eqref{eq:FullDynCont}, given by
\begin{align}
  \label{eq:discretized-dynamics}
  \bar{x}_s=f(\bar{x}_{s-1})+g(\bar{x}_{s-1})\Delta a_{s-1}  
\end{align}
with $\bar{x}_0 = x_0 \equiv x(t)$. For example, $\bar{x}_{3}=f(f(f(\bar{x}_0)+g(\bar{x}_0)\Delta a_0)+g(\bar{x}_1)\Delta a_1)+g(\bar{x}_2)\Delta a_2$. We denote this recursive mapping from $\bar{x}_0$ to $\bar{x}_s$ by $F$, $\bar{x}_s=F(\bar{x}_0; \Delta\control_{0}^{s-1} )$. Then the sensitivity of the state at the time step  $s$ to the action at the  time step $r$ can be calculated via the iterated differentiation chain rule applied to the state derivative of the dynamics $F$:
\begin{align}
  \frac{\partial \bar{x}_{s}}{\partial a_r}=&\prod_{\tau=r+2}^{s}\!\nabla_{\bar{x}} f(\bar{x}_{\tau-1})\;g(\bar{x}_r)\label{eq:LinRes},
\end{align}
where $\nabla_{\bar{x}} f(\bar{x}_{\tau})$ is the  $d_x\times d_x$ Jacobian matrix, which approximates $f$ up to the linear order in the state and  the control. Specifically, the $(i,j)$-th entry of $\nabla_{\bar{x}} f(\bar{x}_{\tau})$ is $\frac{\partial f_i(\bar{x}_{\tau})}{\partial  \bar{x}_{\tau,j}}$, where indices $i,j$ stand for components of the vectors $x$ and $f$.  For $s=r+1$, the expression in \eqref{eq:LinRes} evaluates to $\frac{\partial \bar{x}_{r+1}}{\partial a_r}=g(x_r)$. 

Now we  define the linear response of the sequence of the system's states $x_{s_1}^{s_2}$ to the sequence of the agent's actions $\Delta a_{r_1}^{r_2}$
\begin{equation}\label{eq:LinRespMat}
\mathcal{F}_{r_1, r_2}^{s_1, s_2}({x}_0) =\begin{tikzpicture}[
    node distance=1mm and 0mm,
    baseline]
\matrix (M1) [matrix of nodes,{left delimiter=[},{right delimiter=]}]
{
  $\frac{\partial \bar{x}_{s_2}}{\partial a_{r_2}}$ & $\frac{\partial \bar{x}_{s_2}}{\partial a_{r_2-1}}$ & $\dots$ & $\frac{\partial \bar{x}_{s_2}}{\partial a_{r_1}}$ \\
  $\frac{\partial \bar{x}_{s_2-1}}{\partial a_{r_2}}$ &  $\frac{\partial \bar{x}_{s_2-1}}{\partial a_{r_2-1}}$ & $\dots$  & $\frac{\partial \bar{x}_{s_2-1}}{\partial a_{r_1}}$ \\
  $\vdots$ & $\vdots$ & $\dots$  & $\vdots$ \\
  $\frac{\partial \bar{x}_{s_1}}{\partial a_{r_2}}$ & $\frac{\partial \bar{x}_{s_1}}{\partial a_{r_2-1}}$ & $\dots$ & $\frac{\partial \bar{x}_{s_1}}{\partial a_{r_1}}$\\
};
\draw[{ForestGreen!50},very thick]  (M1-1-1.north west) -| (M1-4-1.south east) -| (M1-1-1.north west);
\draw[{NavyBlue!70},very thick]  (M1-1-4.north west) -| (M1-1-4.south east) -| (M1-1-4.north west);
\draw[Red!80,very thick, dashed]  (M1-1-1.north west) -| (M1-1-4.south east) -| (M1-1-1.north west);
\end{tikzpicture}_{d_x\cdot s\times d_a\cdot r},
\end{equation}
where $s=s_2-s_1+1$, $r=r_2-r_1+1$, $s+\timegapint+r-1=\totaltimeint$, and the entries are computed via \eqref{eq:LinRes}. Usually we consider situations where the agent applies its controls for $r$ time steps, and then after a gap observes the state for $s$ steps. That is, $s_1 = r_2+1 + \timegapint$, where $\timegapint\geq 0$ is the gap between the end of the control sequence and the start of the observations. 

Notice that traditional definitions of sensitivity of a dynamical system to its controls are blocks $\mathcal{F}_{r'_1, r'_2}^{s'_1, s'_2}$ in this overall sensitivity matrix, $\mathcal{F}_{r_1, r_2}^{s_1, s_2}$. For example, if  $r'_1=r'_2=0$, $\timegapint'=\totaltimeint-1$, and $s'_1=\totaltimeint$, then  $s'_2=\totaltimeint$, and the sensitivity matrix collapses to just the entries that measure the sensitivity of the current state to the controls  during the immediately preceding time step, $\mathcal{F}_{0, 0}^{\totaltimeint, \totaltimeint}(x_0)=\frac{\partial \bar{x}_{\totaltimeint}}{\partial  a_0}$. This is also the  blue block of the overall sensitivity matrix,  \eqref{eq:LinRespMat}. 

With the definitions above, in the linear response regime, the effect of a sequence of (small) actions on a sequence of states, \eqref{eq:FullDynCont}, becomes
\begin{align}
    \Delta x_{s_1}^{s_2} = \mathcal{F}_{r_1, r_2}^{s_1, s_2}(x_0) \Delta a_{r_1}^{r_2} + \tilde{\eta},\label{eq:LinResp_Noise}
\end{align}
%\stnote{not sure how to align with \eqref{eq:FullDyn}, where both $\Delta a$ and $\Delta \eta$..}
where $\Delta a$ and $\Delta x$ are the reverse-time-ordered vectors of small actions and the induced deviations of states (which themselves can be vectors).%\footnote{We use $a$ or $A$ for generic action sequences. However, when actions are small, we use $\Delta a$ or $\Delta A$ to emphasize the smallness.} 
Here $\tilde{\eta}$ models both the total noise resulting from the integration of the process noise $d\eta$ from \eqref{eq:FullDynCont} and the noise of the subsequent observation of the state perturbation $\Delta x_{s_1}^{s_2}$, which we assume as Gaussian.

\subsubsection*{Generalized  Empowerment} 
Since the entire dynamics is now linear, cf.~\eqref{eq:LinResp_Noise}, we can consider  formally effects of arbitrary length sequences of actions on arbitrary length sequences of future states. In other words, we can  define the {\em  generalized empowerment},
\begin{align}
     \C^{\totaltimeint,\actiontimeint,\timegapint}(\state_0)
  &\defineas \underset{p(\vec{\control}\mid
    \state_0)}{\max}\;\;\I(X_{\actiontimeint+\timegapint}^{\totaltimeint};
    A_{0}^{\actiontimeint-1}| \state_0)\label{setEmpMaxGeneral}\;.
\end{align}
%{\color{Red}
%Daniel's comment:
%1. Gap: 0 to \totaltimeint-\timegapint-1\\
%2. Gap: \totaltimeint-\timegapint-1 to \totaltimeint-\timegapint-1 + dt\\
%3. Gap: \totaltimeint-\timegapint-1+dt to \totaltimeint
%}
Here, $\actiontimeint$ denotes the number of time steps at which
actions are performed, $\timegapint$ is the time gap between the
action sequence and the beginning of the observation of the resulting
states, and $\totaltimeint$ is the last step in that observed
sequence. That is,  $\C^{\totaltimeint,\actiontimeint,\timegapint}$ measures the maximum mutual information between a sequence of actions and a later sequence of states, rather than just one final state, like empowerment does. 

Plugging in \eqref{eq:LinResp_Noise} into \eqref{EmpOptAction}, we observe that  computing the generalized empowerment in discretized time with an arbitrary discretization step and an arbitrary time horizon $T_e$ reduces to a traditional calculation of the  channel capacity of a linear Gaussian channel, though with a large number of dimensions reflecting both the duration of the signal and the duration of the response.  Specifically, 
\begin{align}
  \C^{\totaltimeint,\actiontimeint,\timegapint}(x_0)=&\underset{\substack{\channelpower_{i}\ge 0\\{\textstyle\sum\limits_i}\channelpower_i = P}}{\max} \;\frac{1}{2}\sum_{i=1}^{d_x}\ln(1+\rho_{i}(x_0)\channelpower_{i})\label{WaterFilling}.
\end{align}
Here $\rho_{i}(x_0)$ are the singular values of the appropriate submatrix 
$\mathcal{F}_{r'_1, r'_2}^{s'_1, s'_2}(x_0)$; for example, the traditional empowerment corresponds to the red-dashed submatrix in \eqref{eq:LinRespMat}. Further, $P$ is the {\em power} of the control signal $\Delta a$ over the whole control period, and $\channelpower_i\geq 0$ is that part of the overall power of the control signal which is associated with the $i$-th singular value (called {\em channel power}). The channel power can be computed by the usual {\em water-filling} procedure \cite{Cover}. Note that here we denote $P$ as power, as per control-theoretic convention, but since we fix the time interval over which it is applied, the  units of $P$  are those  of energy. As per our {\em weak control} assumption, we assume $P$ to be suitably small.

With \eqref{WaterFilling}, calculation of any generalized empowerment becomes tractable, at least in principle. This also shows explicitly that the (generalized) empowerment is a function of the sensitivity matrix ${\mathcal F}$, and with it of quantities used to characterize dynamics, such as the Lyapunov exponents.

To compute $\C^{\totaltimeint,\actiontimeint,\timegapint}(x_0)$ efficiently for an arbitrary dynamical system \eqref{eq:FullDynCont} and arbitrary long time horizons and arbitrary small discretization steps, we start by discretizing the time and calculating the linear response matrix ${\cal F}$. While in this paper we do this by analytical differentiation, numerical differentiation can be used whenever $f$ is unknown. We then calculate the singular values of ${\cal F}$; this is straightforward on modern computers for dimensionalities of up to a few hundred.  Finally, we apply the ``water filling'' procedure to find the set of channel powers $\channelpower_i$ to match the available total power $P$ in % and vary the height of the filling line to match the constraint on power to solve
\eqref{WaterFilling}, and from there we calculate the (generalized) empowerment value. We will employ this approach for all  examples in this paper.

\subsubsection*{Connecting Generalized Empowerment to Related Quantities}

Generalized empowerment with different durations of action and observation sequences is related to various quantities describing  dynamical systems, including those  defining intrinsic motivation \cite{schmidhuber2010formal,wissner2013causal,EmpBerkly1,sharma2019dynamics}. For example,  Causal Entropic Forcing (CEF) \cite{wissner2013causal} is defined as actions that maximize the entropy of future trajectories of a system.  With  $\actiontimeint=1$ and $\timegapint=0$, $\C^{\totaltimeint,\actiontimeint,\timegapint}$ in \eqref{setEmpMaxGeneral} measures the immediate consequences of a single action on a trajectory with  a fixed time horizon $T_e$. Maximizing $\C^{\totaltimeint,\actiontimeint,\timegapint}$ is then equivalent to choosing actions that maximize {\em susceptibility}, and not the entropy of trajectories with a given time horizon. In other words, one can interpret   $\C^{\totaltimeint,1,0}$ as a ``kicked'', or agent-controllable, version of CEF, where just the first action can be selected by the agent at any time, and uncontrolled future variability is discarded in action planning (see Fig.~\ref{fig:scheme} for an illustration). Such kicked CEF corresponds to the green submatrix in \eqref{eq:LinRespMat}.

Now consider the top right corner (blue) of \eqref{eq:LinRespMat} with $T_e=T_a=1$, or, equivalently, $s'_2=s_2$ and $s'_1= s'_2-1$. In the limit of a very long horizon, $s_2\to\infty$, the appropriate submatrix of ${\mathcal F}$ is
\begin{align}
  \Lambda \equiv \underset{s_2\rightarrow \infty}{\lim}\Biggl(\Bigl(\frac{\partial \bar{x}_{s_2}}{\partial a_{r_1}}\Bigr)\Bigl(\frac{\partial \bar{x}_{s_2}}{\partial a_{r_1}}\Bigr)^{\transpose}\Biggr)^{\frac{1}{s_2}},\label{eq:LyapExp}
\end{align}
where ${\transpose}$ is the transpose, and $\frac{\partial \bar{x}_{s_2}}{\partial a_{r_1}}$ is given by \eqref{eq:LinRes}. In the special case that the control gain is the identity, $g(x)=x$, the logarithm of the eigenvalues of $\Lambda$ reduces to the usual characteristic Lyapunov exponents of the dynamical system \cite{abarbanel1992local}. However, once a more general control gain is applied, the action-controlled perturbation, $a_{r_1}$ may be able to affect only  a part of the state space. This means that $\Lambda$ not only is a generalized empowerment with specific indices, but it is also a specialization of the concept  of Lyapunov exponents to the controllable subspace. Thus we refer to the log-spectrum of $\Lambda$ as the {\em control Lyapunov exponents}, cf.~Fig.~\ref{fig:scheme}.

In summary, \eqref{setEmpMaxGeneral} and  the linearization, \eqref{eq:LinRespMat}, provide a unified view of various sensitivties of the dynamics to the controls, and hence on various versions of  intrinsic motivation. 

\begin{figure*}[t!]
  \begin{minipage}{.33\textwidth}
    \includegraphics[width=\linewidth]{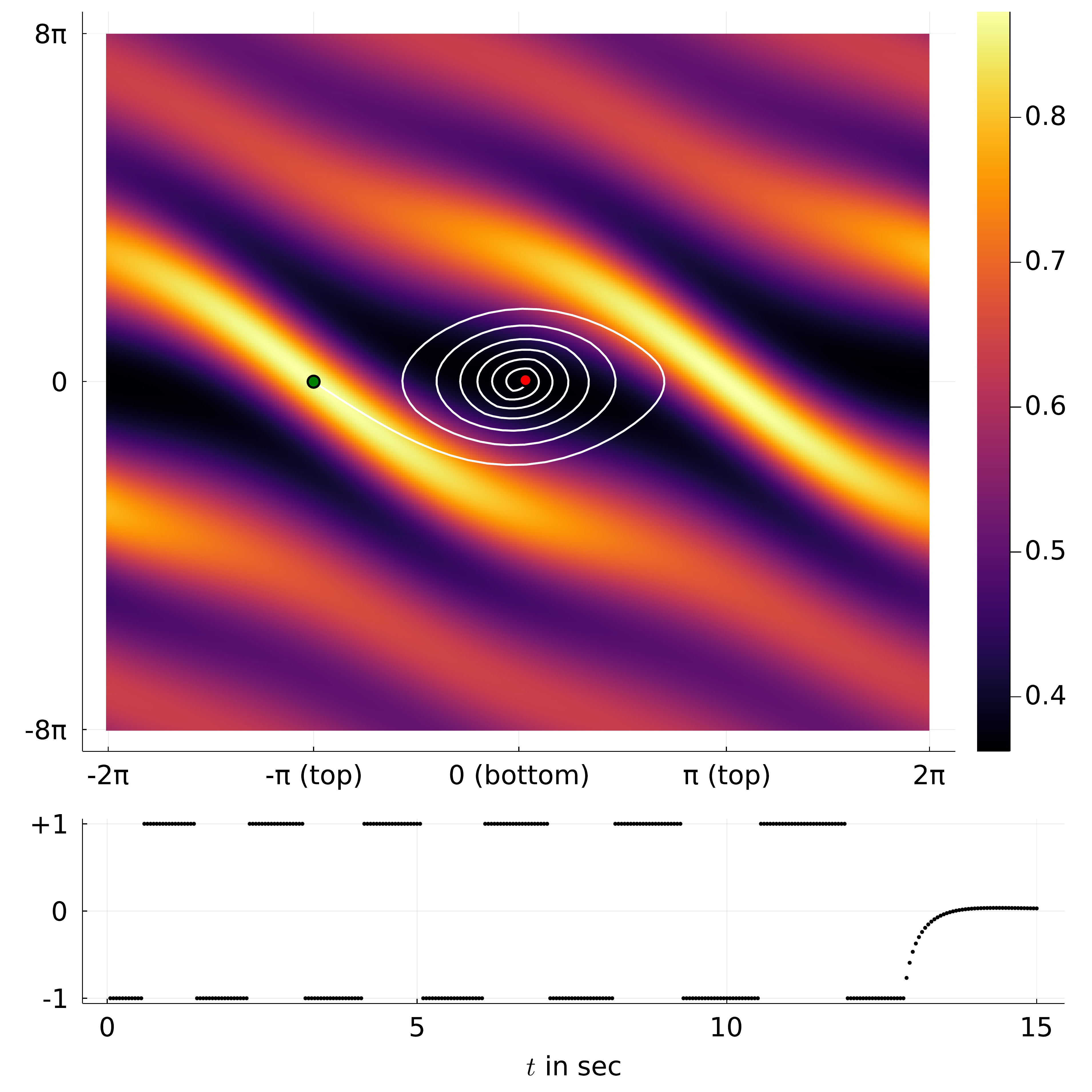}
  \end{minipage}% \quad
  \begin{minipage}{.33\textwidth}
    \includegraphics[width=\linewidth]{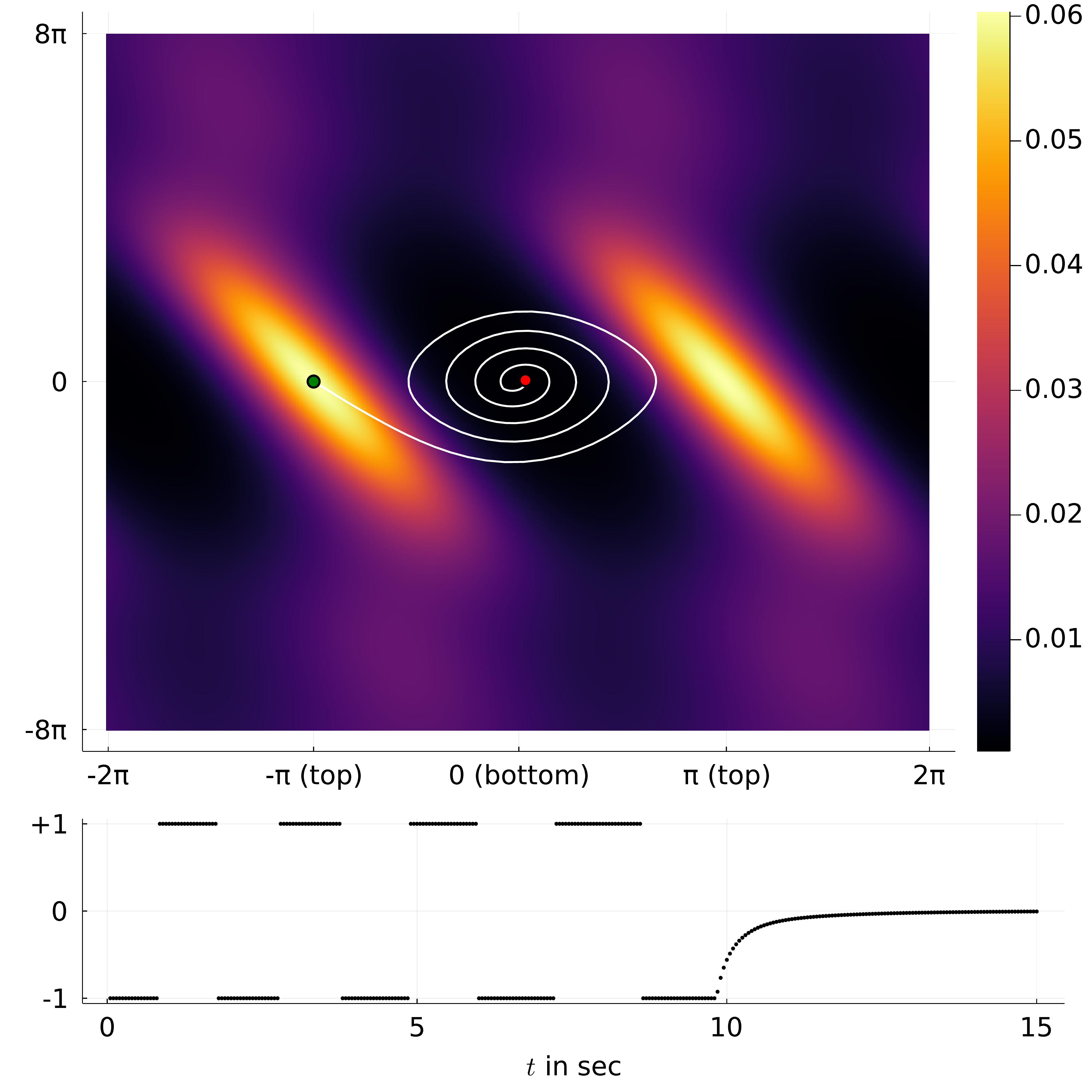}
  \end{minipage}% \quad
  \begin{minipage}{.33\textwidth}
    \includegraphics[width=\linewidth]{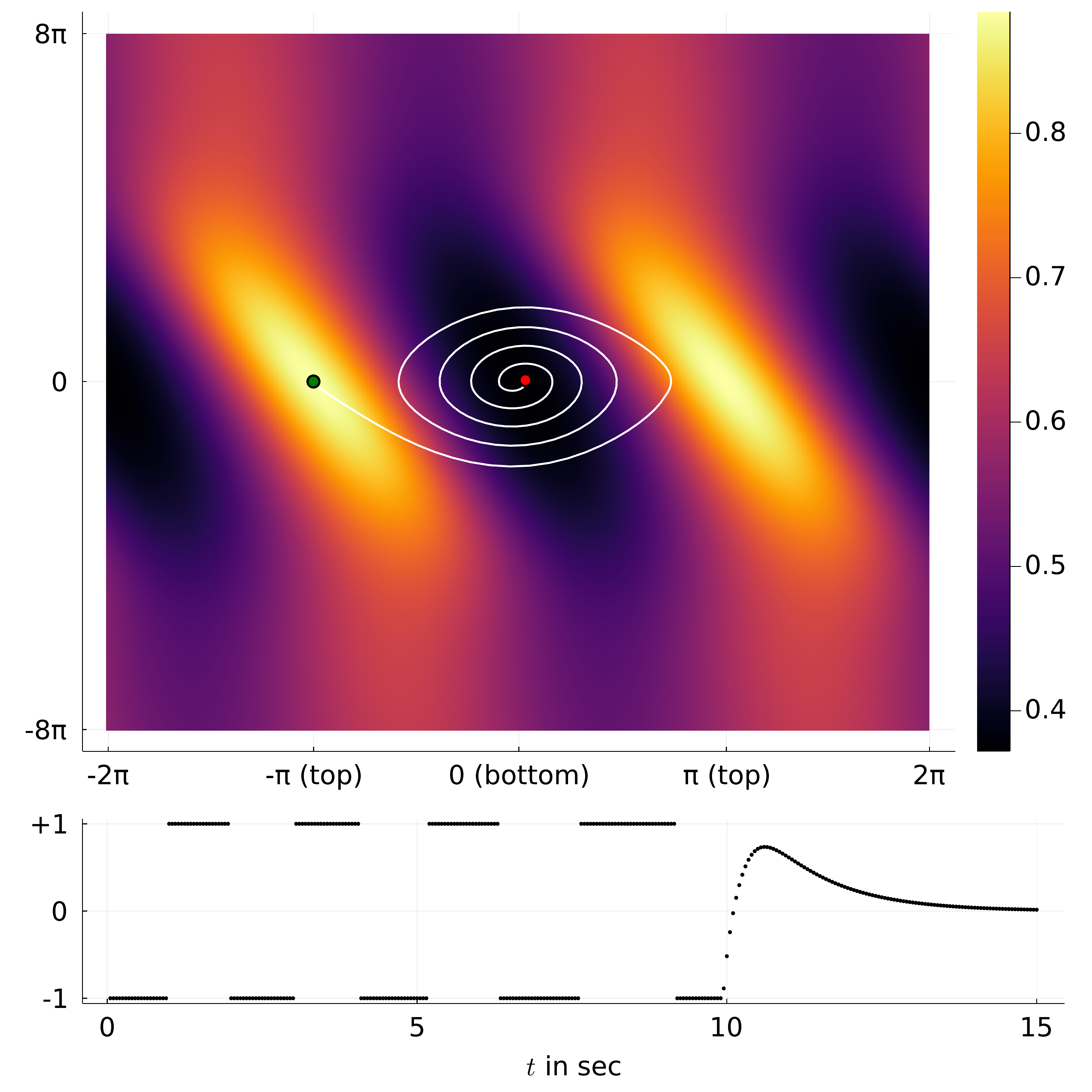}
  \end{minipage}% \quad
  \caption{\label{fig:emp_cef_ale}Intrinsic motivation based control in the power-constrained regime.  Top row: generalized empowerment landscapes in the linear approximation for empowerment (left), controlled Lyapunov exponent (middle), and  kicked CEF (right) versions of the problem, plotted against $\theta$ (horizontal axis) and $\dot{\theta}$ (vertical axis), measured in \si{\radian} and \si{\radian/\second}, respectively. Black dots in each panel are the final state, and white lines are the trajectories of the pendulum, starting at the bottom denoted by the red dots. Bottom row: the control signals chosen from the generalized empowerment maximization as a function of time. Here the time horizon is $t_e=0.5\si{\second}$.
  }
\end{figure*}

\subsection{Intrinsic motivation in power-constrained agents}
An agent controlling a system with unconstrained actions can trivially reach any state in a controllable dynamical system \cite{EVANS} by simply forcing their desired outcome without sophisticated control. Thus to render the setup interesting, we consider only power-constrained, or {\em weak} agents. To show that empowerment maximization, in the linearized regime, is an efficient control principle,  we use it to stabilize a family of inverted pendula (single pole, double pole, and cart-pole), which are simple, paradigmatic models of important phenomena, such as human walking \cite{kuo2007six}.

Solutions for the stabilization problem are known. They require to accumulate energy by swinging the pendulum back and forth into resonance without overshooting and then to keep the pendulum upright. When details of the system are not specified {\em a priori}, this solution needs to be learned by the agent. Finding such an indirect control policy by traditional reinforcement learning is nontrivial \cite{doya2000reinforcement}, since the increasing oscillations require a long time for the balancing to take place, and the acquisition of informative rewards indicating success is significantly delayed.  As we will show, it is precisely in such situations that intrinsic motivation based on empowerment is especially useful, since it is determined from only comparatively local properties of the dynamics along the present trajectory and its potential future variations.

\subsubsection*{Inverted pendulum}
We start with a relatively simple task of swinging up and stabilizing an inverted pendulum without an  external reward. With an angle of $\pendangle$ (in radians)  from the upright vertical, the equations of motion of the pendulum are
\begin{align}
  \label{eq:pendulum-equation-ito}
  \begin{pmatrix}
    d\pendangle(t)\\
    d\angularvelocity(t)
  \end{pmatrix}
  & =
    \begin{pmatrix}
      \angularvelocity(t) dt\\
      \frac{g}{\pendulumlength}\sin( \pendangle(t))\,dt + \frac{d\torque(t)}{m\pendulumlength^2} + \frac{dW(t)}{m\pendulumlength^2}
    \end{pmatrix},
\end{align}
where $\angularvelocity$ is the angular velocity of the pendulum, $m$ is its mass, $\pendulumlength$ is the  length, $\torque$ is the torque applied by the agent, $g$ is the free fall acceleration, and $dW(t)$ is a Wiener process.

We apply a (stochastically chosen) control signal $\torque(t)$ for the duration $\totaltimeint$ and observe the final state $\observedpendangle = \pendangle + \observationnoise$, where $\observationnoise$ is the standard Gaussian observation noise at the final state. Empowerment is then given by the maximally achievable mutual information between $\torque(t)$ and $\observedpendangle$ at a given power level for $\torque(t)$, i.e., the channel capacity between the two.

The observation noise effectively determines the resolution, at which the end state is considered. Note that in our linear approximation the process noise $dW(t)$ undergoes the same gain sequence as the control signal, and thus it rescales the empowerment landscape and changes the behavior of the system. Thus to compare empowerment values in different states, it is essential to include the observation noise.

We now apply our empowerment-based control protocol,  \eqref{EmpOptAction}, to the inverted pendulum.  We calculate the empowerment landscape by using the time-discretized version of Eqs.~(\ref{eq:FullDynCont},  \ref{eq:pendulum-equation-ito}). For this, we map the deterministic part of the dynamics ($f,g$ in \eqref{eq:FullDynCont}) onto discrete time as per \eqref{eq:discretized-dynamics}. We then compute the channel capacity by applying  \eqref{WaterFilling} using the singular values from \eqref{eq:LinResp_Noise}, where states are given by $(\pendangle,\dot{\pendangle})\in \statevalueset$, and actions consist of applying a torque $\torque$. The landscapes for  the original empowerment, the  controlled Lyapunov exponent, and the kicked CEF versions of the problem, all with the time horizons of $\totaltime=\SI{0.5}{\second}$ and the discretization $\Delta t=10^{-3}$ are shown in Fig.~\ref{fig:emp_cef_ale}.
\begin{figure}[b!]
  \centering
    \includegraphics[scale=0.4]{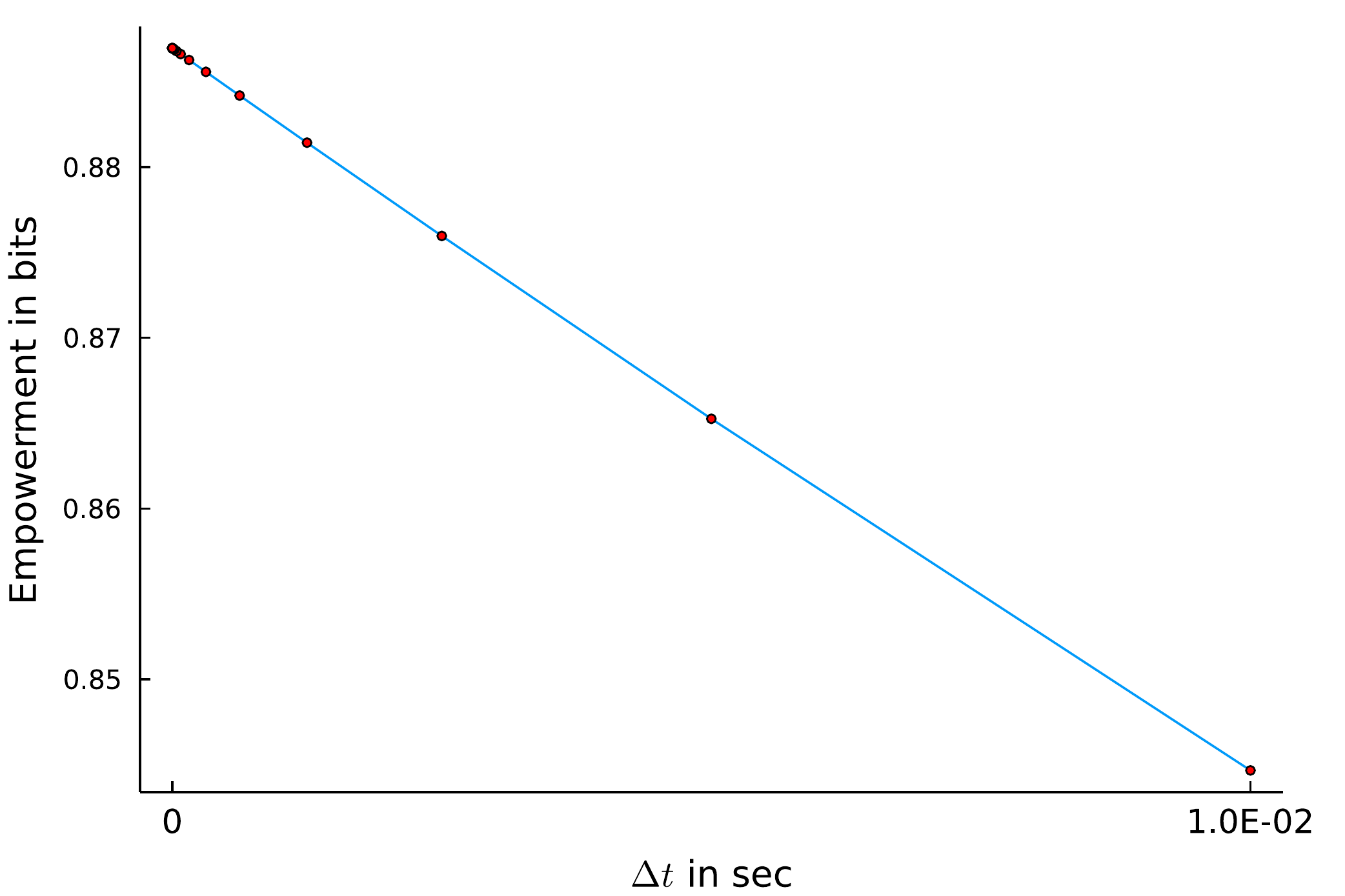}
    \caption{Convergence of the method for $\Delta t \to 0$ and $t_e=0.5\si{\second}$. As time resolution is refined fourfold at every stage, one arrives at a  well-defined value for the empowerment estimation as $\Delta t\to0$. The numerical stability of this limit approximation is consistent throughout the landscape.}
\label{fig:approximation-of-discretized-empowerment}
\end{figure}
Then, from each state, we choose the control action to greedily optimize the generalized  empowerment. The panels in the upper row in this Figure also show trajectories obtained this way. The lower row shows time traces of the control signal derived from the generalized empowerment maximization. In all cases, initially, the agent drives the pendulum at the maximum allowable torque, which we set to be power-constrained to $\SI{\pm 1}{\newton\meter}$. Around 13, 10, and 10 seconds after the start (for the three versions of the empowerment, respectively), the pendulum accumulates enough energy to reach the vertical, and the agents reduce the torques to very  small values, $\torque\ll \SI{1}{\newton\meter}$, which are now sufficient to keep the pendulum in the upright position and prevent it from falling. It is striking that the generalized empowerment landscapes and their induced trajectories are qualitatively similar to those that would be generated by an optimal value function, derived by standard optimal control techniques based on a reward specifically designed to achieve the top position \cite{doya2000reinforcement}. 

In our analysis, we chose a particular discretization  $\Delta t=10^{-3}$ s, and we need to show that our results depend only weekly on this choice. For this, we repeat our analysis at different $\Delta t$. Figure~\ref{fig:approximation-of-discretized-empowerment} shows the dependence of the maximum value of the original empowerment (black dot in left panel of Fig.~\ref{fig:emp_cef_ale}) on $\Delta t$. To the extent that the estimate  converges to a well-defined number linearly as $\Delta t\to 0$, the discrete time  dynamics provides a  consistent approximation to the continuous time dynamics.   

\subsubsection*{Double Pendulum}
\label{sec:double-pendulum}
\begin{figure*}[t!]
  \begin{minipage}{.5\textwidth}
  \centering
    \includegraphics[scale=0.2]{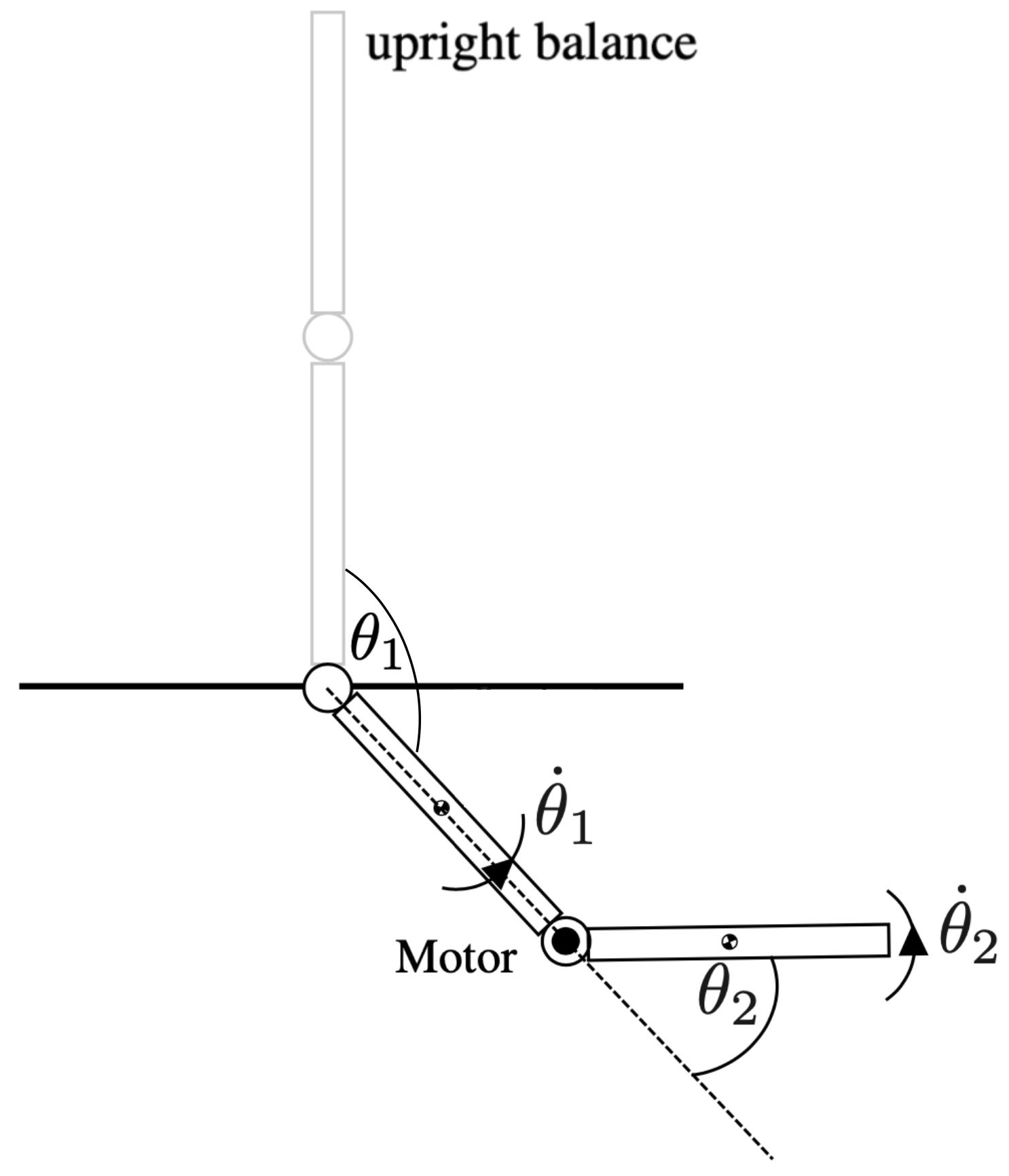}
  \end{minipage}% \quad
  \begin{minipage}{.5\textwidth}
   \centering
    \includegraphics[scale=0.375]{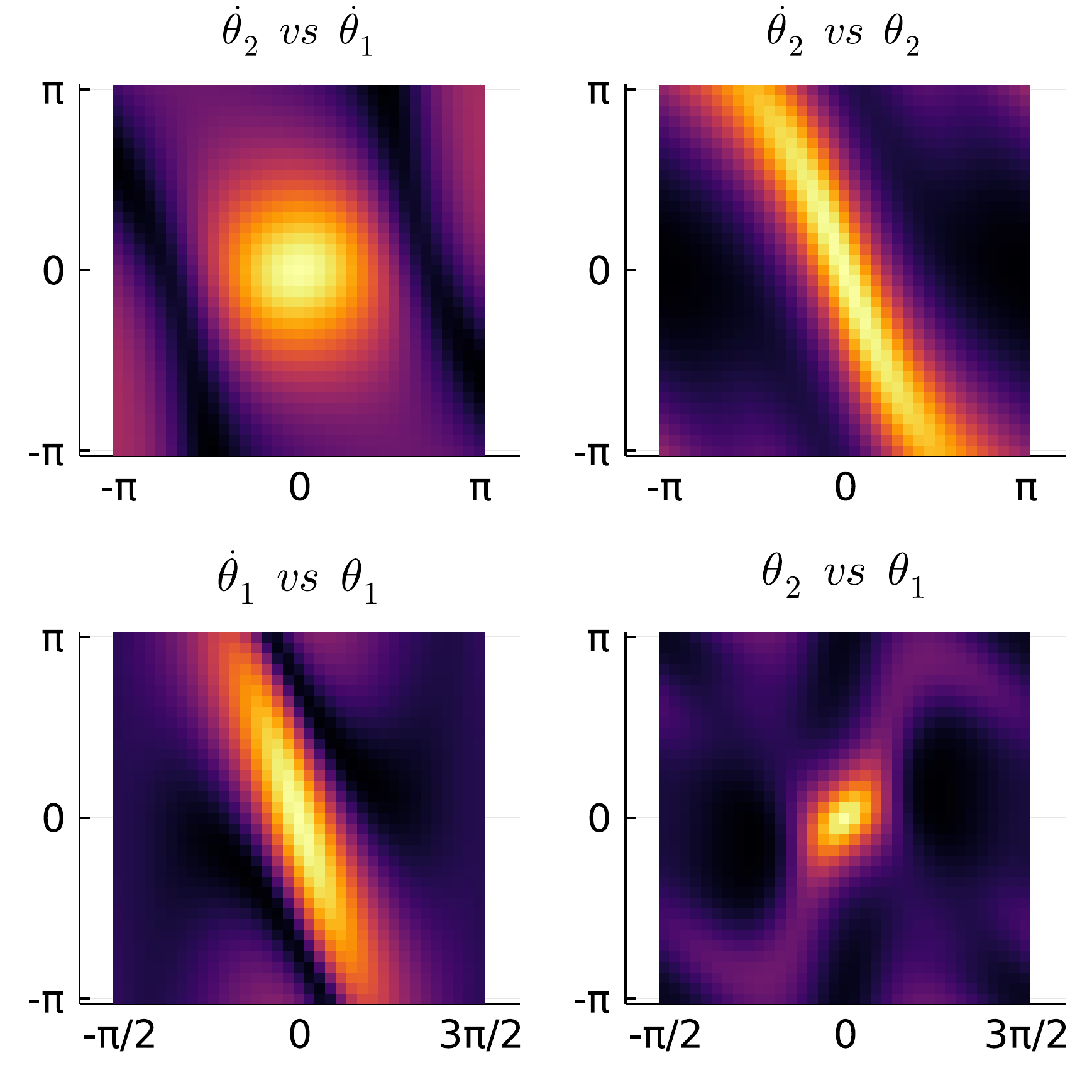}
  \end{minipage}\vspace{-0.275cm}\\
  \begin{minipage}{\textwidth}
 \includegraphics[width=0.975\textwidth]{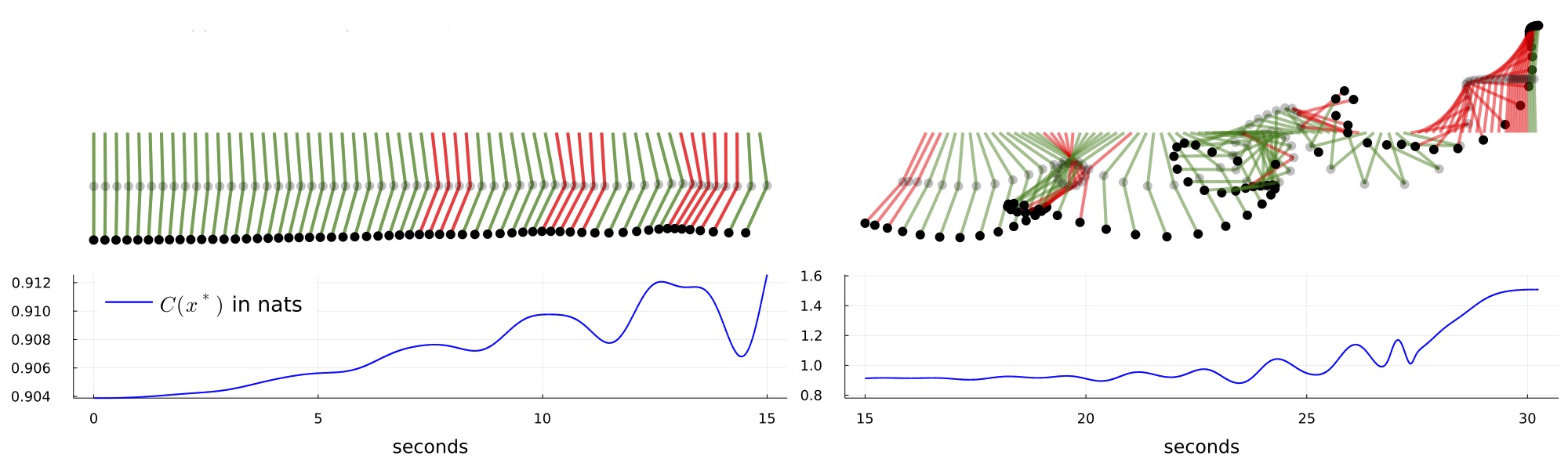}
  \end{minipage}
  \caption{{\bf Top left:} Double pendulum with control torque on the joint between the links with dynamics given by \eqref{eq:double-pend} {\bf Top right:} Slices through the empowerment landscape of a double pendulum. Each subplot shows a particular slice in the 4D landscape, when two other coordinates are zero. For example, the plot with axes $\dot{\theta}_2$,  $\dot{\theta}_1$ is shown for $\theta_2=\SI{0}{rad}$ and $\theta_1=\SI{0}{rad}$. Bottom: Traversing the state space of the double pendulum according to \eqref{EmpOptAction}. The first and the second 15s are shown with different scale for the instantaneous empowerment. The initial and the final positions are both links down and both links up, respectively. Torque is applied to the middle joint only.}\label{fig:double-pendulum}
\end{figure*}

Now we  show that the empowerment maximization formalism is capable of dealing with more challenging problems, such as a power-constrained control of a (potentially chaotic) double pendulum \cite{jung2011empowerment},  Fig.~\ref{fig:double-pendulum}, with equations of motion:
\begin{align}
d\ddot{\theta}_1(t) =& -\frac{1}{d_1(t)}\biggl(d_2(t)\ddot{\theta}_2(t)+\phi_1(t)\biggr)\label{eq:double-pend},\\
d\ddot{\theta}_2(t) =& \frac{1}{m_2\ell_{c_2}^2+I_2-\frac{d^2_2(t)}{d_1(t)}}\biggl(da(t) + dW(t) + \frac{d^2_2(t)}{d_1(t)}\phi_1(t) \nonumber\\
&\qquad\qquad\qquad\quad -m_2\ell_1\ell_{c_2}\dot{\theta}_1(t)^2\sin\theta_2(t) -\phi_2(t)\biggr)\nonumber,
\end{align}
with
\begin{align*}
d_1(t)=&m_1\ell_{c_1}^2+m_2(\ell_1^2 + \ell_{c_2}^2+ 2\ell_1\ell_{c_2}\cos \theta_2(t)) + I_1 + I_2,\\
d_2(t) =& m_2(\ell_{c_2}^2 + \ell_1\ell_{c_2}\cos \theta_2(t)) + I_2,\\
\phi_1(t)=&-m_2\ell_1\ell_{c_2}\dot{\theta}(t)^2\sin \theta_2(t)\! -\!2m_2\ell_1\ell_{c_2}\dot{\theta}_2(t)\dot{\theta}_1(t)\sin \theta_2(t) \\
&\qquad\qquad\qquad\quad+(m_1\ell_{c_1}+m_2\ell_1)g\cos \theta_1(t)+\phi_2(t),\\
\phi_2(t)=&m_2\ell_{c_2}g\cos (\theta_1(t)+\theta_2(t)).
\end{align*}

We add Wiener noise, $dW(t)$, and permit the controller to apply a scalar control signal $|\torque(t)|\le 1 $, at the joint between the two links. In the equations of motion, $m_i$, $\ell_i$, $\ell_{c_i}$, and $I_i$ stand for  the mass, the length, the length to center of mass, and the moment of inertia of the $i$-th link, $i\in [1, 2]$, respectively. Figure~\ref{fig:double-pendulum} shows the landscape for the original empowerment for selected slices of the phase space. This landscape is  more complex than for the single-pendulum. Nonetheless it retains the property that, following the local gradient in the state space directly, one ultimately reaches the state of the maximum empowerment, which is precisely where both links of the pendulum are  balanced upright. The vertical position, however, is a priori not sufficient to  guarantee the balancing since the control only applies torque at the joint
linking the pendulum halves. That is,  the controller  cannot move the pendulum in arbitrary directions through the state space. Surprisingly, this concern notwithstanding, the algorithm still balances the pendulum, cf.~Fig.~\ref{fig:double-pendulum}.

\subsubsection*{Cart-Pole}
\label{sec:cart-pole}
\begin{figure*}[t!]
  \begin{minipage}{.25\textwidth}
  \centering
    \includegraphics[scale=0.112]{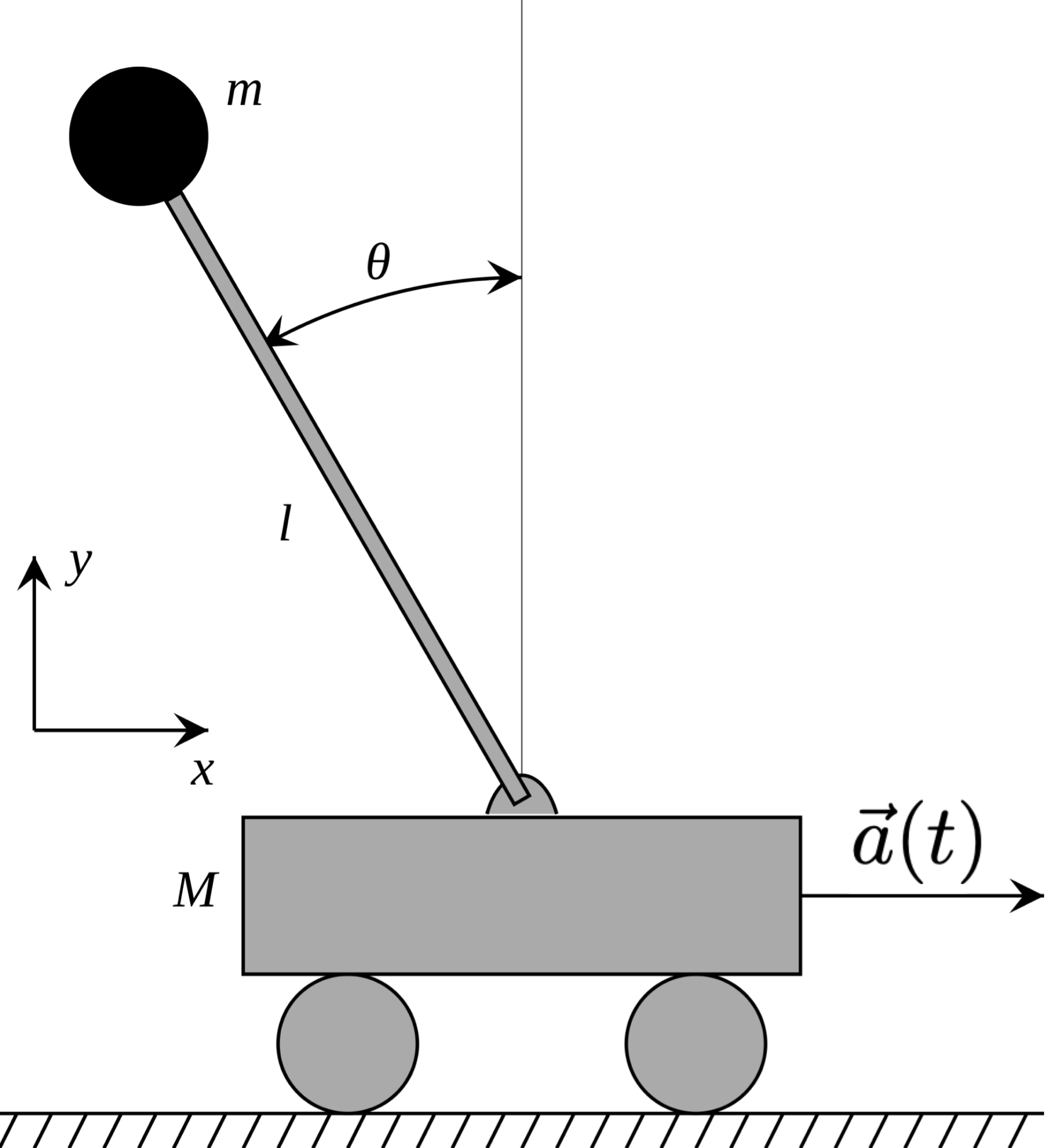}
  \end{minipage}% \quad
  \begin{minipage}{0.75\textwidth}
 \includegraphics[width=0.975\textwidth]{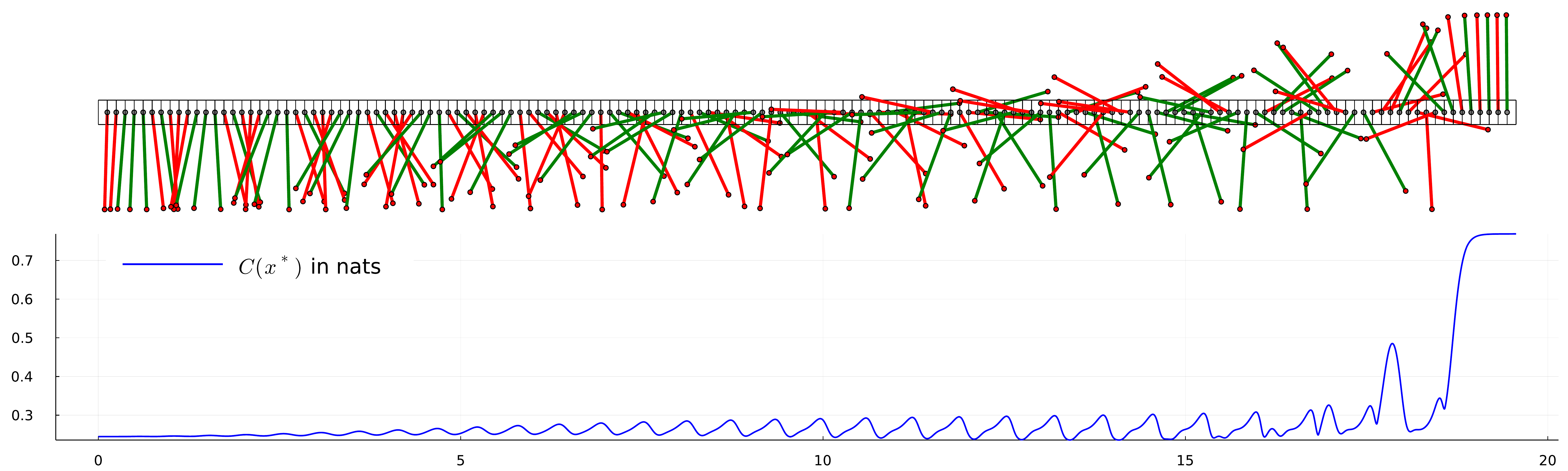}
  \end{minipage}
  \caption{\label{fig:cart-pole}{\bf Left:} Cart-Pole system with control force, $\vec{a}(t)$, applied to the cart only, which moves on the rail (or on the edge of a table), allowing the pole to rotate in the x-y plane. Its dynamics is given by \eqref{eq:cart-pole}. {\bf Right:} Traversing the state space of the pendulum on a cart according to empowerment maximization. The initial and the final state of the  pole are down and up, respectively. The  horizontal axis is time in seconds $t\in [0, 20]s$.}
\end{figure*}
We have additionally verified that the empowerment maximization also balances an inverted pendulum on a moving cart, cf.~Fig.\ref{fig:cart-pole}. Here the control signal (force) is applied to the cart. Thus the pendulum is now affected only indirectly. The dynamics of this system is:
\begin{align}
d\ddot{x}(t) =& \frac{m\sin\theta(t)(\ell\dot{\theta}^2(t)+g\cos \theta(t))+da(t)+dW(t)}{M + m\sin^2\theta(t)}\label{eq:cart-pole},\\
d\ddot{\theta}(t) =& -da(t)\cos \theta(t) - m\ell\dot{\theta}^2(t) \cos\theta(t)\sin\theta(t)\nonumber\\
&\qquad\qquad\qquad\qquad\qquad\qquad\quad-(M+m)g\sin\theta(t)\nonumber,
\end{align}
where $x(t)$, $\theta(t)$, $m$, $M$, $\ell$, $g$, $|a(t)|\le 1$ are the $x$ coordinate of the center of mass of the cart, the angle of the pole, the pole mass, the cart mass, the pole length, the free fall acceleration, and the force applied to the cart.

\section*{Discussion}
\label{sec:conclusion}

In this study, we focused on a class of intrinsic motivation models that mimic decision-making abilities of biological organisms in various situations without explicit reward signals. We used an information-theoretic formulation in which the controller starts with knowledge of the (stochastic) dynamical equations describing the agent and the environment, and then selects actions that ``empower'' the agent. That is, the controller improves its ability to affect the system in the future, as measured by the mutual information between the action sequence and the subsequent responses. This leads the system to the most sensitive points in the state space, which we showed solves a problem known to be difficult for simple reinforcement learning algorithms: balancing inverted pendula. Depending on which subsets of the past actions and future responses are used to drive the intrinsic motivation, our approach interpolates between the original formulation of empowerment maximization, maximization of the ``kicked'' version of Causal Entropic Forcing, and maximization of the ``controlled'' subset of the Lyapunov exponents of the agent-environment pair. This provides insight into which properties of the dynamical system are responsible for the behaviors produced by these different motivation functions.

One big challenge in using information-theoretic quantities is computing them, which can be difficult to do either analytically or from data. Our paper makes a significant contribution to solving this problem in the context of empowerment by providing an explicit algorithm for computing various versions of empowerment, for arbitrary lengths of pasts and futures, using the small noise/small control approximation to the dynamics, while still treating the dynamics as nonlinear. This is often the most interesting regime, modeling weak, power-constrained controllers. Crucially, our algorithm is local, so that climbing up the empowerment gradient only requires estimation of the dynamics in the vicinity of the current state of the system. This should be possible in real control applications by using the data directly, possibly with the help of deep neural networks to approximate the relevant dynamical landscapes~\cite{daniels2015automated,brunton2016discovering,chen2022automated}. Therefore, knowing the exact form of the dynamical system, which could be a potential limitation of our approach, is not strictly required. This opens up opportunities for scaling our method to more complex scenarios.

Our  work suggests that, in addition to the Lyapunov spectrum, defined via the trajectory divergence in time due to a small {\em arbitrary} perturbation, one may want to consider the {\em optimal} Lyapunov spectrum, where the initial perturbation is {\em optimally} aligned with the controllable directions in the dynamics. We defer a systematic study of optimal Lyapunov spectra to future work.  

A potential extension of our analysis relates to social interactions. Interacting agents have their own intrinsic motivations and affect each other's ability to achieve their goals. Understanding how multiple agents interact, each trying to empower itself in the presence of others, and whether and when this leads to cooperation or conflict is a promising area for future research. Crucially, the ability to affect someone else's empowerment may provide insight into what distinguishes social interactions from purely physical interactions among nearby individuals.
\subsection*{Acknowledgements}
ST was supported in part by California State University, and the College of Engineering at SJSU. IN was supported in part by the Simons Foundation Investigator award, the Simons-Emory Consortium on Motor Control, and NIH grant 2R01NS084844. DP acknowledges partial support by the EC H2020-641321 socSMCs FET Proactive project and the Pazy Foundation.

\bibliographystyle{unsrt}
\setstretch{0.8}
\bibliography{ref_short}

\begin{thebibliography}{10}

\bibitem{oudeyer2009intrinsic}
P.-Y. Oudeyer and F.~Kaplan.
\newblock What is intrinsic motivation? a typology of computational approaches.
\newblock {\em Frontiers in neurorobotics}, 1:6, 2009.

\bibitem{sutton2018reinforcement}
R.~S. Sutton and A.~G. Barto.
\newblock {\em Reinforcement learning: An introduction}.
\newblock MIT press, 2018.

\bibitem{doya2000reinforcement}
K.~Doya.
\newblock Reinforcement learning in continuous time and space.
\newblock {\em Neural computation}, 12(1):219--245, 2000.

\bibitem{mohamed2015variational}
S.~Mohamed and D.~J. Rezende.
\newblock Variational information maximisation for intrinsically motivated
  reinforcement learning.
\newblock In {\em Advances in neural information processing systems}, pages
  2125--2133, 2015.

\bibitem{gregor2016variational}
K.~Gregor, D~J. Rezende, and D.~Wierstra.
\newblock Variational intrinsic control.
\newblock {\em arXiv preprint arXiv:1611.07507}, 2016.

\bibitem{baumli2021relative}
K.~Baumli, D.\ Warde-Farley, S.~Hansen, and V.~Mnih.
\newblock Relative variational intrinsic control.
\newblock {\em Proceedings of the AAAI Conference on Artificial Intelligence},
  35(8):6732--6740, 2021.

\bibitem{kwon2020variational}
T.~Kwon.
\newblock Variational intrinsic control revisited.
\newblock {\em International Conference on Learning Representations, (ICLR),
  2021}, 2021.

\bibitem{sharma2019dynamics}
A.~Sharma, S.~Gu, S~Levine, V.~Kumar, and K.~Hausman.
\newblock Dynamics-aware unsupervised discovery of skills.
\newblock {\em arXiv preprint arXiv:1907.01657}, 2019.

\bibitem{sharma2020emergent}
A.~Sharma, M~Ahn, S.~Levine, V.~Kumar, K.~Hausman, and S.~Gu.
\newblock Emergent real-world robotic skills via unsupervised off-policy
  reinforcement learning.
\newblock {\em RSS, Robotics: Science and Systems 2020}, 2020.

\bibitem{eysenbach2018diversity}
B.~Eysenbach, A~Gupta, J.~Ibarz, and S.~Levine.
\newblock Diversity is all you need: Learning skills without a reward function.
\newblock {\em International Conference on Learning Representations, (ICLR),
  2019}, 2018.

\bibitem{houthooft2016vime}
R.~Houthooft, X.~Chen, Y.~Duan, J.~Schulman, F.~De~Turck, and P.~Abbeel.
\newblock Vime: Variational information maximizing exploration.
\newblock {\em Advances in neural information processing systems}, 29, 2016.

\bibitem{achiam2018variational}
J.~Achiam, H.~Edwards, D.~Amodei, and P.~Abbeel.
\newblock Variational option discovery algorithms.
\newblock {\em arXiv preprint arXiv:1807.10299}, 2018.

\bibitem{choi2021variational}
J.~Choi, A.~Sharma, H.~Lee, S.~Levine, and S.~S. Gu.
\newblock Variational empowerment as representation learning for
  goal-conditioned reinforcement learning.
\newblock {\em International Conference on Machine Learning}, pages 1953--1963,
  2021.

\bibitem{salge2014empowerment}
C.~Salge, C.~Glackin, and D.~Polani.
\newblock Empowerment--an introduction.
\newblock In {\em Guided Self-Organization: Inception}, pages 67--114.
  Springer, 2014.

\bibitem{klyubin2005empowerment}
A.~S. Klyubin, D~Polani, and C.~L. Nehaniv.
\newblock Empowerment: A universal agent-centric measure of control.
\newblock {\em IEEE Congress on Evolutionary Computation, 2005}, 1:128--135,
  2005.

\bibitem{jung2011empowerment}
T.~Jung, D.~Polani, and P.~Stone.
\newblock Empowerment for continuous agent—environment systems.
\newblock {\em Adaptive Behavior}, 19(1):16--39, 2011.

\bibitem{klyubin08:_keep_your_option_open}
A.~S. Klyubin, D.~Polani, and C.~L. Nehaniv.
\newblock Keep your options open: An information-based driving principle for
  sensorimotor systems.
\newblock {\em PLoS ONE}, 3(12):e4018, Dec 2008.

\bibitem{zhao2020efficient}
R.~Zhao, P.~Abbeel, and S.~Tiomkin.
\newblock Efficient empowerment estimation for unsupervised stabilization.
\newblock {\em International Conference Learning Representations}, 2020.

\bibitem{zhao2019dynamical}
R.~Zhao, S.~Tiomkin, and P.~Abbeel.
\newblock Dynamical system embedding for efficient intrinsically motivated
  artificial agents.
\newblock {\em Advances in Neural Information Processing Systems, (NeurIPS),
  DeepRL}, 2019.

\bibitem{wissner2013causal}
A.~D. Wissner-Gross and C.~E. Freer.
\newblock Causal entropic forces.
\newblock {\em Physical review letters}, 110(16):168702, 2013.

\bibitem{Cover}
T.~M. Cover and J.~A. Thomas.
\newblock {\em Elements of information theory}.
\newblock John Wiley \& Sons, 2012.

\bibitem{Empowerment1}
A.~S. Klyubin, D.~Polani, and C.~L. Nehaniv.
\newblock Empowerment: A universal agent-centric measure of control.
\newblock In {\em 2005 IEEE Congress on Evolutionary Computation}, volume~1,
  pages 128--135. IEEE, 2005.

\bibitem{EmpBerkly1}
B.~Eysenbach, A.~Gupta, J.~Ibarz, and S.~Levine.
\newblock Diversity is all you need: Learning skills without a reward function.
\newblock In {\em ICLR}, 2019.

\bibitem{Du2020AvEAV}
Y.~Du, S.~Tiomkin, E.~Kiciman, D.~Polani, P.~Abbeel, and A.~D. Dragan.
\newblock Ave: Assistance via empowerment.
\newblock {\em Neural Information Processing Systems, NeurIPS 2020}, 2020.

\bibitem{salge2017empowerment}
C.~Salge and D.~Polani.
\newblock Empowerment as replacement for the three laws of robotics.
\newblock {\em Frontiers in Robotics and AI}, 4:25, 2017.

\bibitem{bialek2001predictability}
W.~Bialek, I.~Nemenman, and N.~Tishby.
\newblock Predictability, complexity, and learning.
\newblock {\em Neural computation}, 13(11):2409--2463, 2001.

\bibitem{holmes}
C.~Holmes and I.~Nemenman.
\newblock Estimation of mutual information for real-valued data with error bars
  and controlled bias.
\newblock {\em Phys Rev E}, 100:022404, 2019.

\bibitem{Empowerment9}
T.~Jung, D.~Polani, and P.~Stone.
\newblock Empowerment for continuous agent—environment systems.
\newblock {\em Adaptive Behavior}, 19(1):16--39, 2011.

\bibitem{anonymous2021efficient}
R.~Zhao, K.~Lu, P.~Abbeel, and S.~Tiomkin.
\newblock Efficient empowerment estimation for unsupervised stabilization.
\newblock In {\em International Conference on Learning Representations, ICLR},
  2021.

\bibitem{CEF}
A.~D. Wissner-Gross and C.~E. Freer.
\newblock Causal entropic forces.
\newblock {\em Physical review letters}, 110(16):168702, 2013.

\bibitem{ng1999policy}
A.~Y. Ng, D.~Harada, and S.~Russell.
\newblock Policy invariance under reward transformations: Theory and
  application to reward shaping.

\bibitem{dayan1993feudal}
P.~Dayan and G.~E. Hinton.
\newblock Feudal reinforcement learning.
\newblock In {\em Advances in neural information processing systems}, pages
  271--278, 1993.

\bibitem{hrl2016}
T.~D. Kulkarni, K.~Narasimhan, A.~Saeedi, and J.~Tenenbaum.
\newblock Hierarchical deep reinforcement learning: Integrating temporal
  abstraction and intrinsic motivation.
\newblock In {\em Advances in Neural Information Processing Systems, NeurIPS},
  2016.

\bibitem{pathakICLR19largescale}
Y.~Burda, H.~Edwards, D.~Pathak, A.~Storkey, T.~Darrell, and A.~A. Efros.
\newblock Large-scale study of curiosity-driven learning.
\newblock In {\em ICLR}, 2019.

\bibitem{pathakICMl17curiosity}
D.~Pathak, P.~Agrawal, A.~A. Efros, and T.~Darrell.
\newblock Curiosity-driven exploration by self-supervised prediction.
\newblock In {\em ICML}, 2017.

\bibitem{Volks}
M.~Karl, M.~Soelch, P.~Becker-Ehmck, D.~Benbouzid, P.~van~der Smagt, and
  J.~Bayer.
\newblock Unsupervised real-time control through variational empowerment.
\newblock 2018.

\bibitem{Empowerment2}
C.~Salge, C.~Glackin, and D.~Polani.
\newblock Approximation of empowerment in the continuous domain.
\newblock {\em Advances in Complex Systems}, 16(02n03):1250079, 2013.

\bibitem{Empowerment3}
C.~Salge, C.~Glackin, and D.~Polani.
\newblock Empowerment--an introduction.
\newblock In {\em Guided Self-Organization: Inception}, pages 67--114.
  Springer, 2014.

\bibitem{Empowerment12}
T.~Anthony, D.~Polani, and C.~L. Nehaniv.
\newblock General self-motivation and strategy identification: Case studies
  based on sokoban and pac-man.
\newblock {\em IEEE Transactions on Computational Intelligence and AI in
  Games}, 6(1):1--17, 2014.

\bibitem{charlesworth2019intrinsically}
H.~J. Charlesworth and M.~S. Turner.
\newblock Intrinsically motivated collective motion.
\newblock {\em Proceedings of the National Academy of Sciences},
  116(31):15362--15367, 2019.

\bibitem{schmidhuber2010formal}
J.~Schmidhuber.
\newblock Formal theory of creativity, fun, and intrinsic motivation
  (1990--2010).
\newblock {\em IEEE Transactions on Autonomous Mental Development},
  2(3):230--247, 2010.

\bibitem{abarbanel1992local}
H.~D. Abarbanel, R.~Brown, and M.~B. Kennel.
\newblock Local lyapunov exponents computed from observed data.
\newblock {\em Journal of Nonlinear Science}, 2(3):343--365, 1992.

\bibitem{EVANS}
L.~C. Evans.
\newblock An introduction to mathematical optimal control theory version 0.2.
\newblock 1983.

\bibitem{kuo2007six}
A.~D. Kuo.
\newblock The six determinants of gait and the inverted pendulum analogy: A
  dynamic walking perspective.
\newblock {\em Human movement science}, 26(4):617--656, 2007.

\bibitem{daniels2015automated}
B.~C. Daniels and I.~Nemenman.
\newblock Automated adaptive inference of phenomenological dynamical models.
\newblock {\em Nature communications}, 6(1):1--8, 2015.

\bibitem{brunton2016discovering}
S.~L. Brunton, J.~L. Proctor, and J.~N. Kutz.
\newblock Discovering governing equations from data by sparse identification of
  nonlinear dynamical systems.
\newblock {\em Proceedings of the national academy of sciences},
  113(15):3932--3937, 2016.

\bibitem{chen2022automated}
B.~Chen, K.~Huang, S.~Raghupathi, I.~Chandratreya, Q.~Du, and H.~Lipson.
\newblock Automated discovery of fundamental variables hidden in experimental
  data.
\newblock {\em Nature Computational Science}, 2(7):433--442, 2022.

\end{thebibliography}

\end{document}